\documentclass[letterpaper]{article} 
\usepackage[preprint]{aaai2027}  
\usepackage[hyphens]{url}  
\usepackage{graphicx} 
\usepackage{natbib}  
\usepackage{caption} 
\usepackage{dblfloatfix}
\usepackage{algorithm}
\usepackage{algorithmic}
\usepackage{booktabs}
\usepackage{amsmath,amssymb}

\newcommand{\method}{PAGE-RAG}

\title{PAGE-RAG: Provenance-Aware Graph Evidence Promotion for Fixed-Budget Multi-hop Retrieval-Augmented Generation}
\author{
    Haokun Deng\equalcontrib\textsuperscript{\rm 1},
    Xunkai Li\equalcontrib\textsuperscript{\rm 1},
    Hongchao Qin\textsuperscript{\rm 1},
    Rong-Hua Li\textsuperscript{\rm 1}\thanks{Correspondence to: Rong-Hua Li \textless{}lironghuabit@126.com\textgreater{}.}
}
\affiliations{
    \textsuperscript{\rm 1}Department of Computer Science, Beijing Institute of Technology\\
    Beijing, China
}

\begin{document}
\maketitle

\begin{abstract}
Multi-hop question answering in retrieval-augmented generation (RAG) often benefits from retrieving beyond the few candidates that will finally be read: narrow retrieval can miss an indispensable hop, while expanded retrieval introduces topical distractors. This challenge is not tied to a particular knowledge-base format. Candidate pools may come from standalone retrievers, standard RAG backends, or graph-based retrieval pipelines. What is needed is a query-aware selection layer that can use relational structure to filter candidates before generation. PAGE-RAG addresses this setting by using a graph as a temporary selection structure, rather than assuming a graph-structured knowledge base. It builds a query-local graph over retrieved candidates, records why candidates are connected, and treats each connection as a support hypothesis rather than support itself. We identify the resulting failure mode as a \emph{connectivity-support gap}: connected candidates do not necessarily support the answer. We propose PAGE-RAG, a \textbf{P}rovenance-\textbf{A}ware \textbf{G}raph \textbf{E}vidence promotion method that scores candidate paths with relevance, source-tracing metadata, specificity, hubness, noise, and coherence signals, and applies minimal sufficient selection to promote supporting facts into a compact reader context. PAGE-RAG can serve as a complete retrieval-to-reading pipeline, and the same promotion stage can be inserted after existing retrieval or RAG systems without replacing their upstream retrieval logic. Across three multi-hop QA benchmarks under the same final budget, PAGE-RAG improves support F1 and answer F1 by 10.4 and 3.3 points on a weighted average over a strong retriever. As a plug-in, PAGE-RAG further improves all reported RAG backends, including reasoning-oriented, compression-based, graph-based, and document/chunk-level systems.
\end{abstract}

\section{Introduction}

Retrieval-augmented generation (RAG) helps language models answer knowledge-intensive questions by retrieving external context before generation \citep{guu2020realm,lewis2020rag,izacard2021fid,izacard2022atlas}. Graph-based RAG further organizes retrieved candidates through entities, relations, or passage links, allowing the system to use relationships among candidate texts rather than treating each passage independently \citep{edge2024graphrag,gutierrez2024hipporag,guo2025lightrag,zhu2025kg2rag}. Even with these mechanisms, multi-hop question answering remains challenging: a faithful answer requires complementary facts to be retrieved together and placed in the right relation \citep{yang2018hotpotqa,ho2020twoWiki,trivedi2022musique,trivedi2023ircot}.

A narrow top-\emph{k} retrieval may miss an indispensable hop, making the answer unreachable \citep{trivedi2023ircot,zhuang2024efficientrag,liu2026opera}. A natural remedy is to retrieve more candidates, but the enlarged pool also brings more topical distractors \citep{xu2023recomp,hwang2025exit,asai2024selfrag}. In graph-based retrieval, the problem becomes sharper: shared entities and relation links can create plausible but non-supporting bridges, and these connected distractors can degrade the final answer \citep{gutierrez2025hipporag2,guo2025lightrag,zhu2025kg2rag,luo2025gfmrag}. The central challenge is to expose enough candidate facts while preventing noisy or merely connected candidates from dominating the final context.

\begin{figure*}[t]
\centering
\includegraphics[width=0.98\textwidth]{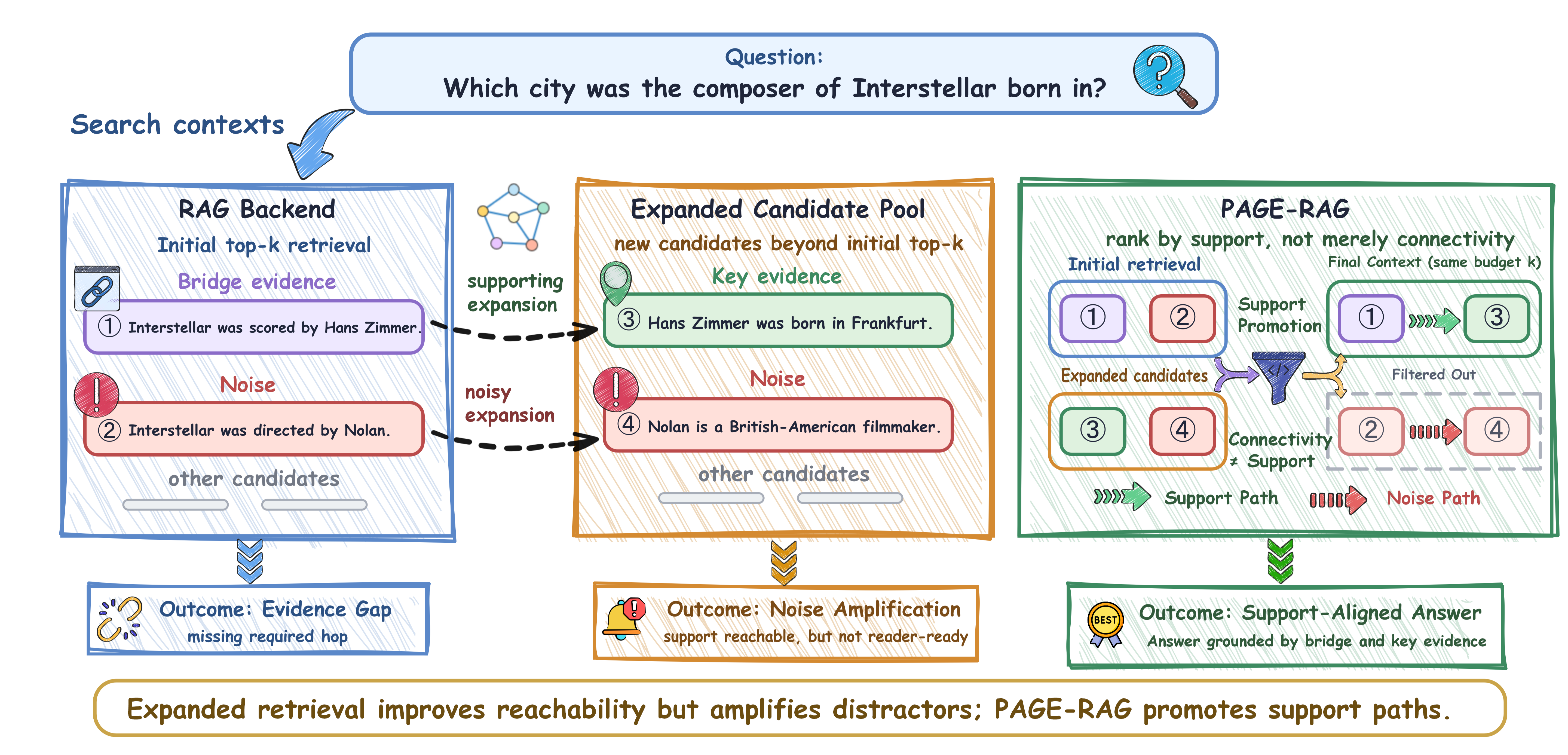}
\caption{Motivating example for fixed-budget support promotion. Expanded retrieval can recover a missing supporting fact, but it can also amplify distractors. \method\ promotes support-bearing candidates into the same reader budget.}
\label{fig:intro-motivation}
\end{figure*}

Figure~\ref{fig:intro-motivation} illustrates this failure mode. For the question ``Which city was the composer of \emph{Interstellar} born in?'', the initial retrieval finds a useful bridge fact about Hans Zimmer but misses the birthplace fact; it also retrieves a director-related distractor. Expanding the pool recovers the key fact, but it can also extend the distractor path through Christopher Nolan. The example exposes the core gap: \textbf{connectivity is not support}. Connectivity explains why candidates are related, while support explains why they can justify the answer. The two are not inherently equivalent.

To bridge this gap, we propose \method, a provenance-aware graph evidence promotion method for multi-hop RAG. Given a question and an expanded candidate pool from an upstream retriever, \method\ builds a query-local support graph over the retrieved candidates. It treats graph edges as support hypotheses rather than answer support by default, scores candidate paths with query alignment, source reliability, bridge specificity, hubness, noise, and path coherence signals, and then applies minimal sufficient selection to produce a compact final context. In this way, \method\ preserves the reachability gained from expanded retrieval while avoiding the mistake of optimizing raw relevance or connectivity.

\method\ can operate as a complete retrieval-to-reading pipeline, and its promotion stage can also be attached after existing retrievers, non-graph RAG systems, or graph-based RAG systems that expose candidate pools. This plug-in interface is lightweight: it does not require replacing the upstream system, changing the reader, or rebuilding a corpus-wide graph. Instead, \method\ uses a temporary query-local graph to decide which retrieved connections should be promoted into the final context.

Our contributions are summarized as follows.

\begin{enumerate}
\item \textbf{Candidate-Pool Quality Gap.} We identify a quality risk in RAG knowledge construction: expanded candidate pools may contain the needed facts, yet their topical or graph-based relations can still fail to jointly support the answer. We name this broader issue the candidate-pool quality gap, with the connectivity-support gap as its graph-side manifestation, and frame support promotion as a meaningful research direction for multi-hop RAG.
\item \textbf{Flexible Plug-in Promotion.} We introduce \method\ as a backend-general promotion layer that can be inserted after retrievers, standard RAG backends, or graph-based RAG systems. It builds a temporary query-local graph over their candidate pools, scores support hypotheses from connection metadata, and selects a compact context without replacing the upstream system.
\item \textbf{Complete PAGE-RAG Pipeline.} We instantiate the same idea as a full retrieval-to-reading pipeline with expanded retrieval, query-local graph construction, support-aware path scoring, and minimal sufficient selection, separating useful support from raw relevance or connectivity before generation.
\item \textbf{Empirical Validation.} We show that \method\ improves support F1 and answer F1 by 10.4 and 3.3 points on a weighted average over a strong retriever, and brings consistent gains when used as a plug-in for existing RAG systems, measured by answer quality and support quality.
\end{enumerate}

\section{Preliminaries}

\subsection{Problem Definition}
We study multi-hop RAG with an upstream knowledge access module and a bounded reader context. Given a question and an external corpus or knowledge base, the upstream module returns an expanded candidate pool. The module may be a standalone retriever, a standard RAG backend, or a graph-based retrieval system; our problem does not assume that the underlying knowledge base is itself a graph. Each candidate may be a sentence, passage, document, or chunk, depending on the backend.

The reader receives at most \(k\) final context units. The goal is therefore not to pass the whole expanded pool to the reader, but to select a compact subset that jointly supports the answer. This differs from ordinary top-\(k\) retrieval, which ranks candidates mainly by local relevance, and from graph retrieval, which may treat connected candidates as useful because they are structurally related. We call the broader failure the \emph{candidate-pool quality gap}: an expanded pool may contain the needed facts, yet the final context can still be dominated by related but non-supporting candidates. In graph-structured selection, this appears as the \emph{connectivity-support gap}, where connected candidates do not necessarily support the answer. PAGE-RAG addresses this problem by using a temporary query-local graph to select support-bearing candidates under the same reader budget.

\subsection{Related Work}
\textbf{Retrieval and candidate expansion.}
Standalone retrievers provide candidate texts for open-domain QA and RAG. DPR and Contriever use learned dense representations \citep{karpukhin2020dpr,izacard2022contriever}, ColBERTv2 keeps fine-grained token interaction \citep{santhanam2022colbertv2}, and NV-Embed represents a recent LLM-embedding retriever family \citep{lee2024nvembed}. These methods can expose semantically related candidates, but multi-hop QA often requires complementary facts whose relevance becomes clear only after another hop is found. Expanding retrieval increases the chance of finding the missing hop, while also adding distractors, which motivates a downstream support-promotion step.

\textbf{Reasoning-oriented and selective RAG.}
RAG systems also improve retrieval through reasoning, feedback, or context control. IRCoT and EfficientRAG interleave retrieval with multi-step reasoning or query refinement \citep{trivedi2023ircot,zhuang2024efficientrag}; Self-RAG and corrective retrieval methods use reflection or assessment signals to decide how retrieved content should be used \citep{asai2024selfrag,yan2024crag}; planner-executor and agentic retrieval frameworks decompose complex questions into sub-goals \citep{yao2023react,liu2026opera}. Context compression and selection methods reduce distracting text before reading \citep{xu2023recomp,hwang2025exit,jiang2024longllmlingua}. These methods make RAG more adaptive, but they do not directly model whether the retained candidates form a support-bearing chain for the answer.

\textbf{Graph-based RAG.}
Graph-based RAG organizes text with entities, relations, passages, or knowledge graph structures. Earlier reasoning-path retrieval explores paths over Wikipedia-style graphs \citep{asai2020reasoningpaths}; recent systems include GraphRAG, the HippoRAG series, LightRAG, and KG2RAG \citep{edge2024graphrag,gutierrez2024hipporag,gutierrez2025hipporag2,guo2025lightrag,zhu2025kg2rag}. Graphs help reveal relationships that independent passage scores may miss, but shared entities, generic relations, and noisy extraction can also connect candidates that are related without supporting the answer. PAGE-RAG differs by using graph structure as a query-local selection workspace: edges propose support hypotheses, and connected paths must pass support-aware scoring and minimal selection before entering the reader context.

\section{Method}

\begin{figure*}[t]
\centering
\includegraphics[width=0.98\textwidth]{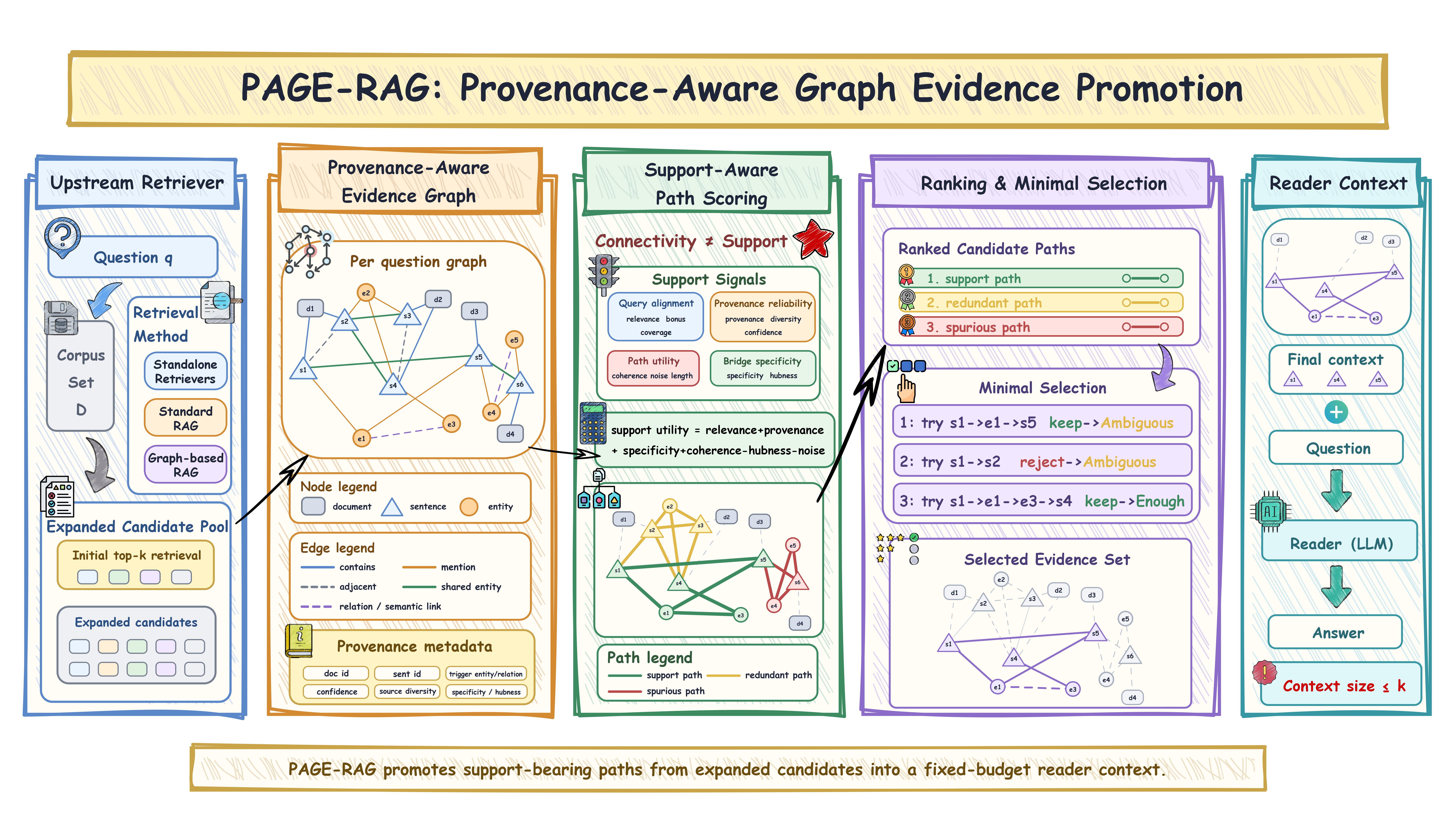}
\caption{Overview of \method. The main pipeline retrieves an expanded candidate pool, builds a per-question provenance-aware graph, scores candidate paths with support signals, performs minimal selection, and passes a fixed-size context to the reader.}
\label{fig:framework}
\end{figure*}

We introduce \method, a retrieval-to-reading framework for multi-hop RAG. In its main pipeline, \method\ starts from a question and a corpus, uses a neural retrieval stage to obtain an expanded candidate pool, builds a query-local provenance-aware graph over these candidates, scores candidate paths by support rather than raw connectivity, and returns a compact reader context. The same promotion stage can also be attached after an existing retriever or RAG backend when that system already exposes a candidate pool. The method follows the principle that \textbf{connectivity is not support}: a graph edge indicates that two candidates may be related, but it does not by itself support the answer. Figure~\ref{fig:framework} summarizes the full workflow.

\subsection{Expanded Candidate Retrieval}

The motivation for the retrieval stage is reachability. In multi-hop QA, an initial top-\(k\) list may contain an obvious bridge fact but miss a later hop that is less lexically similar to the question. Expanding retrieval is a direct way to make such missing facts reachable, but it should create a search space for selection rather than a larger reader input.

Operationally, the first stage of \method\ retrieves a broad candidate pool for the input question. In our main pipeline, this stage uses NV-Embed-v2 as the neural retriever \citep{lee2024nvembed}. Instead of retrieving only the final top-\(k\) items that will be shown to the reader, \method\ retrieves a larger pool. The initial top-\(k\) results are therefore treated as the conventional reader budget, while the remaining candidates provide additional search space where missing support facts may appear.

This expanded pool is not used by simply giving the reader more text. It is passed to the PAGE-RAG promotion stages, which build the local graph, score support paths, and select a final context whose size is no larger than \(k\). The same interface also allows backend generalization: an external retriever, standard RAG system, or graph-based RAG system may provide the expanded pool, and PAGE-RAG can then be inserted between that retrieval stage and the reader to perform support-aware promotion.

\subsection{Provenance-Aware Evidence Graph}

The expanded pool is only a list, so it does not explain how candidates may jointly answer the question. PAGE-RAG introduces a graph at this point to expose relational structure among candidates, but the graph is used as a query-local workspace for selection, not as a claim that connected candidates are already supporting facts. This distinction is important because the same entity link can either connect complementary facts or create a distracting bridge.

\method\ converts the expanded pool into a per-question support graph. The construction is similar in spirit to graph-based RAG systems that organize text through chunks, entities, relations, or passage links \citep{edge2024graphrag,gutierrez2024hipporag,guo2025lightrag,zhu2025kg2rag}. Candidate documents or passages are connected to their sentences, sentences are connected to mentioned entities, neighboring sentences are linked within the same source, and candidates from different sources can be linked through shared entities or extracted relations. This gives PAGE-RAG a local structure for exploring possible multi-hop support without assuming that every connected candidate is useful.

The key difference from a connectivity-only graph is that each edge stores why the connection exists. PAGE-RAG records the following source-tracing metadata for each edge:

\begin{equation}
\label{eq:edge-metadata}
\begin{aligned}
m(e)=\{&\mathrm{src},\mathrm{sent},\mathrm{trig},\mathrm{conf},\\
&\mathrm{div},\mathrm{spec},\mathrm{hub}\}.
\end{aligned}
\end{equation}

Here, source and sentence identifiers make the edge traceable to the original text; the trigger entity or relation explains what created the edge; and confidence and diversity estimate reliability. Specificity and hubness form the key pair for separating useful bridges from broad connectors. For a shared-entity edge with trigger set \(T_e\) and local entity set \(T\), we compute:

\begin{equation}
\label{eq:specificity-hubness}
\begin{aligned}
\mathrm{spec}(e)&=\operatorname{Clamp}\!\left(
\frac{\operatorname{mean}_{t\in T_e}\mathrm{idf}(t)}
{\max_{t'\in T}\mathrm{idf}(t')}\right),\\
\mathrm{hub}(e)&=\operatorname{Clamp}\!\left(
\frac{\operatorname{mean}_{t\in T_e}\mathrm{df}(t)}
{\max_{t'\in T}\mathrm{df}(t')}\right).
\end{aligned}
\end{equation}

Here, \(\mathrm{df}\) and \(\mathrm{idf}\) are computed within the query-local graph rather than the whole corpus. Thus, a rare trigger entity receives a high specificity score, while a trigger that connects many local candidates receives a high hubness penalty.

These metadata also determine how edges are used during path construction. Document--sentence and sentence--entity edges serve mainly as source anchors. Candidate support paths are expanded primarily over sentence-level links, especially neighboring-sentence and shared-entity edges. A path through a specific entity with a reliable source trace can be promoted, whereas a path through a generic hub entity should be downweighted even if it connects many candidates.

This provenance-aware design changes the role of the graph. Many graph-based RAG pipelines use the graph as an indexing or retrieval substrate, where connectivity helps surface related chunks or subgraphs. PAGE-RAG uses the graph as a temporary workspace for evaluating support, not as a corpus-wide knowledge store. The graph is rebuilt around the expanded pool of the current question, and its connections are later tested by support-aware path scoring before entering the reader context.

\subsection{Support-Aware Path Scoring}

Once the local graph is built, the key question becomes which connected paths should be trusted. A path can be useful because it links complementary facts, redundant because it repeats what is already known, or spurious because it passes through a hub entity or weak relation. The scoring module is designed to turn raw connectivity into comparable support hypotheses before any final context is selected.

A candidate path is scored from the sentences it contains, the sentence-level links it traverses, and the source-tracing metadata stored on those links. PAGE-RAG first assigns each edge a support-oriented score:

\begin{equation}
\label{eq:edge-score}
\begin{aligned}
\mathrm{edge}(e)=
&\alpha\,\mathrm{rel}(e,q)+\beta\,\mathrm{reliab}(e)\\
&+\gamma\,\mathrm{spec}(e)-\delta\,\mathrm{hub}(e)\\
&-\eta\,\mathrm{noise}(e).
\end{aligned}
\end{equation}

Here, \(\mathrm{rel}\) measures question-edge lexical alignment, \(\mathrm{reliab}\) combines confidence and source diversity, and \(\mathrm{noise}\) penalizes weak shared-entity links or low-confidence matches. The path score then combines sentence support, edge support, coherence, and length cost:

\begin{equation}
\label{eq:path-score}
\begin{aligned}
\mathrm{score}(\pi)=
&\ \mathrm{sent}(\pi)+\sum_{e\in E_\pi}\mathrm{edge}(e)\\
&+\lambda\,\mathrm{coh}(\pi)-\mu\,\mathrm{len}(\pi).
\end{aligned}
\end{equation}

These expressions are intentionally compact: the important point is not the exact weight of each term, but the separation between being connected and being useful for answering. In Figure~\ref{fig:framework}, the green path links complementary facts and receives high sentence support, source reliability, and specificity; the yellow path is related but largely overlaps with already available information; the red path is connected through a distracting bridge and has high noise or hubness. The scorer converts path structure and source-tracing metadata into comparable support hypotheses, so the following selection stage can reason over ranked candidates rather than raw graph connectivity.

\subsection{Ranking and Minimal Selection}

Scoring alone is not enough because the final reader context is small and redundancy is costly. Even high-scoring paths can overlap with already selected facts, while a lower-ranked path may provide the missing complementary hop. The role of this module is therefore to convert a ranked path list into a minimal set of context units that is sufficient for answering.

After scoring, \method\ ranks candidate paths but does not simply pass the top paths to the reader. This step is central to the fixed-budget setting. The expanded pool is valuable because it increases the chance of exposing a missing hop, but it is also the source of additional noise. Ranking identifies promising support hypotheses; minimal selection decides which of them actually deserve space in the final context.

Let \(S\) denote the context units already selected for the current question. The overall objective is to choose a compact subset that maximizes support utility under the reader budget:

\begin{equation}
\label{eq:minimal-selection-objective}
S^\star =
\arg\max_{S:\, |S|\le k}
\mathrm{Support}(S,q)-\lambda \mathrm{Cost}(S,q).
\end{equation}

For a candidate path \(\pi\), \(\Delta \mathrm{support}(\pi \mid S)\) measures the additional support contributed by adding the path to \(S\). It is estimated from observable signals such as new question coverage, path score, source diversity, relation match, and whether the path connects complementary facts not already covered by \(S\). \(\Delta \mathrm{cost}(\pi \mid S)\) measures the additional reader budget consumed by the path, including newly introduced sentences or documents, redundancy with selected context, and noise penalties. The support state is judged only from the retrieved text and graph-derived features, not from benchmark supporting-fact annotations. A candidate path is kept only when:

\begin{equation}
\label{eq:incremental-selection}
\Delta \mathrm{support}(\pi \mid S) > \Delta \mathrm{cost}(\pi \mid S).
\end{equation}

The selector accumulates context units incrementally. It tries candidate paths in ranked order, keeps a path when it adds new support, rejects paths whose marginal contribution is redundant or noisy, and stops once the selected context is sufficient or the budget is reached. To avoid discarding useful intermediate bridges too early, PAGE-RAG uses an ambiguity-tolerant continuation rule. If the current selected subgraph is still ambiguous but has positive utility, it remains provisional while the selector considers the next ranked path or a one-hop frontier continuation. If this continuation does not improve the support state or utility, the newly considered path is rejected.

In the example in Figure~\ref{fig:framework}, the first support path is kept, but the selected subgraph is still ambiguous because it does not yet contain enough complementary facts. The selector then rejects a non-supporting path and keeps another support path that changes the state to enough. This turns a high-recall but noisy candidate pool into a compact support-bearing context, making the improvement attributable to better support selection rather than to giving the reader more input.

\subsection{Reader Context and Backend Generalization}

The last design goal is interface compatibility. PAGE-RAG should improve the context that reaches the reader without requiring a new reader architecture or a special prompt format. It should also respect the native unit of the upstream backend: some systems expose sentences, while others expose passages, chunks, or full documents.

The final reader context is serialized together with the question and passed to the reader. In sentence-level settings, PAGE-RAG selects sentence-level support paths and returns the supporting sentences that remain after minimal selection. In document or passage-level settings, PAGE-RAG-doc keeps the same internal sentence/path reasoning, but it aggregates support back to the upstream system's native units. That aggregation is not a mere change in output granularity: document-level reranking also incorporates document length, title-question overlap, retriever rank and score, question token and entity coverage, sentence and path support statistics, cross-document path structure, and source-tracing edge signals such as specificity, hubness, and noise. The core principle stays the same, but the context unit becomes the one expected by the backend and the reader.

\method\ is a complete retrieval-to-reading pipeline, not only a plug-in wrapper. In the main pipeline, the retriever discovers a broad candidate pool, PAGE-RAG promotes support-bearing paths, and the reader answers from the selected context. At the same time, the promotion stage is backend-general: it can be attached after existing retrievers, standard RAG systems, or graph-based RAG systems that expose candidate pools. This makes PAGE-RAG easy to integrate into existing retrieval stacks without changing their reader interface or replacing their upstream retrieval logic. Its bounded role remains the same in all settings: it cannot recover facts that never enter the candidate pool, but it can turn an expanded pool into a compact context that is more useful for generation.

\section{Experiments}

\begin{table*}[!t]
\centering
\setlength{\tabcolsep}{3.5pt}
\renewcommand{\arraystretch}{0.92}
\resizebox{\textwidth}{!}{
\begin{tabular}{lrrrrrrrr}
\toprule
\textbf{Method}
& \multicolumn{2}{c}{HotpotQA}
& \multicolumn{2}{c}{MuSiQue}
& \multicolumn{2}{c}{2Wiki}
& \multicolumn{2}{c}{Avg.} \\
\cmidrule(lr){2-3}\cmidrule(lr){4-5}\cmidrule(lr){6-7}\cmidrule(lr){8-9}
& Ans F1 & Sup F1 & Ans F1 & Sup F1 & Ans F1 & Sup F1 & Ans F1 & Sup F1 \\
\midrule
Contriever \citep{izacard2022contriever} & 65.42 & 35.12 & 46.62 & 38.28 & 56.04 & 31.67 & 58.12 & 33.53 \\
ColBERTv2 \citep{santhanam2022colbertv2} & 62.27 & 30.73 & 42.79 & 34.29 & 52.22 & 29.20 & 54.53 & 30.25 \\
NV-Embed-v2 \citep{lee2024nvembed} & 67.80 & 49.03 & 48.79 & 52.39 & 53.99 & 40.96 & 58.00 & 44.86 \\
\midrule
GraphRAG \citep{edge2024graphrag} & 61.79 & 34.88 & 35.47 & 31.65 & 48.73 & 33.35 & 51.62 & 33.67 \\
\midrule
IRCoT@5 \citep{trivedi2023ircot} & 68.47 & 48.87 & 40.76 & 44.40 & 54.00 & 40.40 & 57.36 & 43.64 \\
IRCoT + \method & 70.87 & 54.79 & 55.18 & \textbf{62.73} & \underline{63.13} & \textbf{54.39} & \underline{64.83} & \textbf{55.42} \\
RECOMP@5 \citep{xu2023recomp} & 68.52 & 43.42 & 45.44 & 50.84 & 57.59 & 45.52 & 59.89 & 45.40 \\
RECOMP + \method & \textbf{71.12} & \underline{54.91} & \underline{55.71} & \underline{62.00} & 63.03 & 54.11 & \underline{64.92} & 55.22 \\
\midrule
\method & \underline{70.94} & \textbf{55.09} & \textbf{56.00} & \underline{62.00} & \textbf{63.23} & \underline{54.16} & \textbf{65.00} & \underline{55.31} \\
\bottomrule
\end{tabular}
}
\caption{Sentence-level results under a fixed top-5 sentence budget. Best results are bolded and second-best results are underlined.}
\label{tab:sentence-level-results}
\vspace{0.6em}

\resizebox{\textwidth}{!}{
\begin{tabular}{lrrrrrrrr}
\toprule
\textbf{Method}
& \multicolumn{2}{c}{HotpotQA}
& \multicolumn{2}{c}{MuSiQue}
& \multicolumn{2}{c}{2Wiki}
& \multicolumn{2}{c}{Avg.} \\
\cmidrule(lr){2-3}\cmidrule(lr){4-5}\cmidrule(lr){6-7}\cmidrule(lr){8-9}
& Ans F1 & Sup F1 & Ans F1 & Sup F1 & Ans F1 & Sup F1 & Ans F1 & Sup F1 \\
\midrule
SelfRAG \citep{asai2024selfrag} & 68.46 & 49.20 & 42.08 & 46.47 & 62.71 & 56.08 & 62.39 & 52.77 \\
SelfRAG + PAGE-RAG & 76.62 & 55.13 & 53.75 & 56.02 & 68.08 & 62.74 & 69.36 & 59.50 \\
EfficientRAG \citep{zhuang2024efficientrag} & 66.58 & 46.41 & 47.17 & 20.04 & 47.08 & 40.57 & 60.65 & 42.19 \\
EfficientRAG + PAGE-RAG & 67.82 & 54.00 & 47.37 & 41.15 & 50.90 & 43.51 & 62.25 & 50.52 \\
\midrule
GFM-RAG \citep{luo2025gfmrag} & 73.08 & 50.69 & 36.11 & 37.88 & 71.48 & 59.10 & 60.22 & 49.22 \\
GFM-RAG + PAGE-RAG & 73.74 & 51.54 & 41.27 & 39.87 & 71.84 & 59.20 & 62.28 & 50.20 \\
LightRAG \citep{guo2025lightrag} & 74.80 & 60.08 & 35.87 & 43.96 & 60.48 & 49.58 & 57.05 & 51.21 \\
LightRAG + PAGE-RAG & 74.81 & 61.04 & 37.13 & 45.56 & 61.94 & 51.15 & 57.96 & 52.58 \\
KG2RAG \citep{zhu2025kg2rag} & 71.30 & 50.48 & 43.58 & 48.99 & 56.76 & 40.61 & 57.21 & 46.69 \\
KG2RAG + PAGE-RAG & 71.80 & 50.73 & 50.11 & 51.95 & 59.87 & 41.20 & 60.59 & 47.96 \\
\midrule
HippoRAG2 \citep{gutierrez2025hipporag2} & 72.45 & 52.39 & 39.20 & 43.65 & 52.58 & 45.33 & 57.71 & 47.48 \\
HippoRAG2 + PAGE-RAG & 74.67 & 53.62 & 42.89 & 45.45 & 53.78 & 46.43 & 59.51 & 48.70 \\
\bottomrule
\end{tabular}
}
\caption{Mixed-granularity results under a fixed top-5 document, passage, or chunk budget. Each PAGE-RAG row is compared with its corresponding upstream backend.}
\label{tab:document-level-results}
\end{table*}

Our experiments validate \method\ from two complementary directions. First, we evaluate it as a complete retrieval-to-reading workflow that starts from expanded retrieval and produces its own compact reader context. Second, we evaluate the same support-promotion stage as a plug-in inserted between existing upstream candidate generators and the reader. Together, these settings test whether \method\ improves final-context quality both as a standalone pipeline and as a backend-general selection layer, without increasing the final reader input.

\subsection{Experimental Setup}

We evaluate on three multi-hop QA benchmarks, HotpotQA \citep{yang2018hotpotqa}, MuSiQue \citep{trivedi2022musique}, and 2WikiMultiHopQA \citep{ho2020twoWiki}, and report supporting-fact F1 (Sup F1) and answer F1 (Ans F1). The two metrics are related but not identical: higher support quality does not always translate linearly into answer quality \citep{asai2024selfrag,trivedi2022musique}. Weighted averages are computed by the number of evaluated examples in each dataset.

For answer generation, we use DeepSeek-V4-Pro \citep{deepseek2026v4}. For baselines with their own generator, such as SelfRAG, we keep the retrieval workflow and feed its selected context to the same reader.

We consider two evaluation protocols. The sentence-level protocol fixes the final reader input to the same number of sentences and compares PAGE-RAG with standalone retrievers, a connectivity-based graph baseline, and reasoning or compression-oriented RAG systems. The GraphRAG row uses GraphRAG-style candidate linking as a controlled graph-construction baseline, rather than the original corpus-level summarization workflow. The mixed-granularity protocol covers systems whose native output unit may be a document, passage, or chunk, so Table~\ref{tab:document-level-results} is not intended for global vertical ranking and is best read through paired baseline/plugin comparisons.

For plug-in experiments, the upstream system first produces an expanded candidate pool using its own retrieval or graph construction procedure, and PAGE-RAG operates between this candidate-generation step and reading. To preserve input-budget fairness, all systems pass at most five final units to the reader; if minimal selection returns fewer units, PAGE-RAG fills the remaining slots with the highest-scoring support candidates.

\subsection{Main Results}

We first examine \method\ as a complete sentence-level retrieval-to-reading pipeline in Table~\ref{tab:sentence-level-results}. Compared with standalone retrievers, \method\ improves the average over the strongest retriever, NV-Embed-v2 \citep{lee2024nvembed}, from 58.00 to 65.00 Ans F1 and from 44.86 to 55.31 Sup F1. The gain is also not limited to retrieval-only baselines. \method\ outperforms the controlled IRCoT@5 \citep{trivedi2023ircot} and RECOMP@5 \citep{xu2023recomp} pipelines by 7.64 and 5.11 points in Ans F1, respectively, and by 11.67 and 9.91 points in Sup F1. Overall, \method\ achieves the best answer F1 in the sentence-level table, while also reaching the second-best support F1. This supports the main claim that PAGE-RAG improves retrieval-side context quality and, more importantly, promotes candidates that better support the answer, leading to stronger downstream generation.

We next evaluate the same promotion stage as a plug-in for recent and competitive RAG backends. On MuSiQue, PAGE-RAG improves IRCoT@5 \citep{trivedi2023ircot} by 14.42 Ans F1 and 18.33 Sup F1, and improves RECOMP \citep{xu2023recomp} by 10.27 Ans F1 and 11.16 Sup F1. Table~\ref{tab:document-level-results} further shows gains for all six reported document/chunk-level backends, including SelfRAG \citep{asai2024selfrag}, EfficientRAG \citep{zhuang2024efficientrag}, HippoRAG2 \citep{gutierrez2025hipporag2}, GFM-RAG \citep{luo2025gfmrag}, LightRAG \citep{guo2025lightrag}, and KG2RAG \citep{zhu2025kg2rag}. The improvements across multi-step retrieval, compression, self-reflective RAG, efficient RAG, and graph-based RAG show that PAGE-RAG is robust across diverse systems, rather than only strengthening a simple retriever.

These results also clarify the scope of the method. PAGE-RAG can improve diverse upstream systems without increasing the final reader budget, but it still depends on the expanded candidate pool provided by the upstream retriever or RAG backend. If the required support facts never appear in that pool, the query-local graph cannot recover them by selection alone. The gains on graph-based backends are stable but generally smaller than the gains on several non-graph RAG settings, because graph RAG systems already expose some relational structure before PAGE-RAG is applied. In these cases, PAGE-RAG mainly recalibrates graph connectivity into support-aware selection: graph structure is useful, but connected candidates still need to be filtered by whether they jointly support the answer.

\begin{figure}[t]
\centering
\includegraphics[width=\columnwidth]{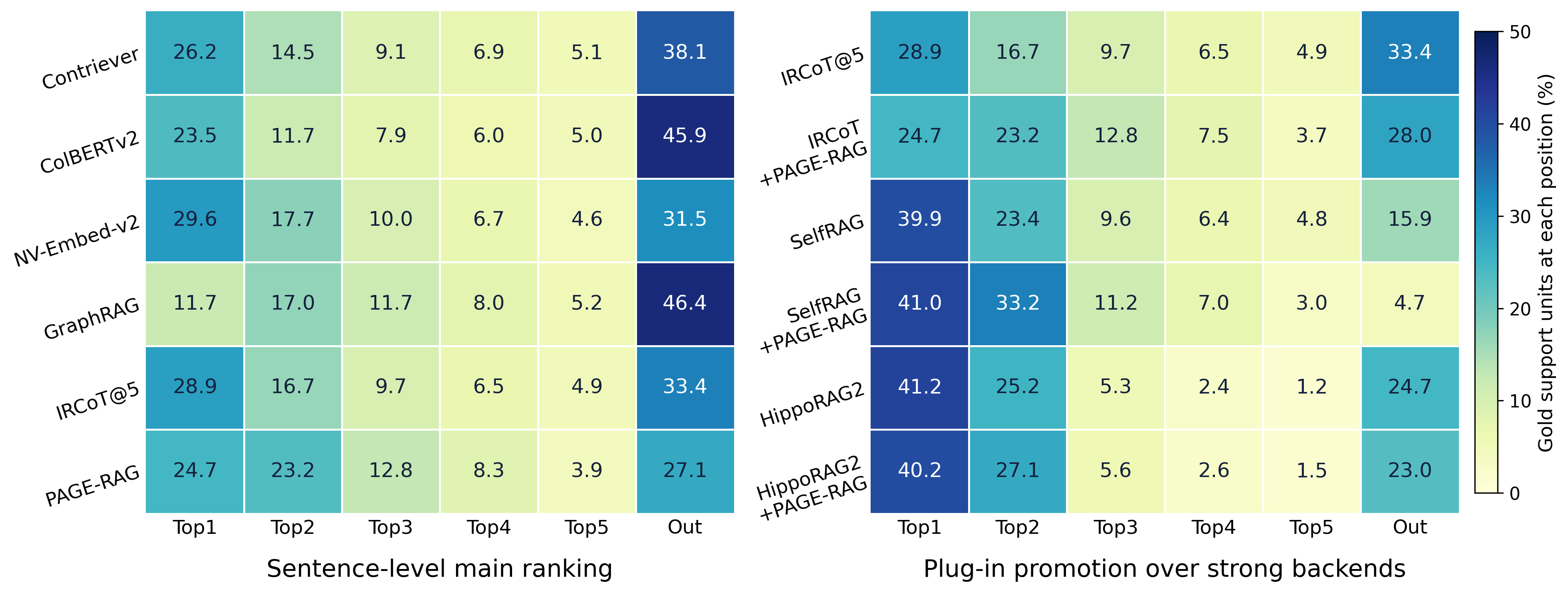}
\caption{Position distribution of gold supporting facts. The left panel compares sentence-level methods, and the right panel compares plug-in pairs.}
\label{fig:support-position-heatmap}
\end{figure}

\begin{table}[t]
\centering
\small
\setlength{\tabcolsep}{4pt}
\begin{tabular}{lcccc}
\toprule
Method & HotpotQA & MuSiQue & 2Wiki & Avg. \\
\midrule
w/o support-aware scoring & 58.10 & 37.54 & 48.76 & 50.64 \\
w/o minimal selection & 61.91 & 39.45 & 52.93 & 54.44 \\
\midrule
\method & \textbf{70.94} & \textbf{56.00} & \textbf{63.23} & \textbf{65.00} \\
\bottomrule
\end{tabular}
\caption{Component ablation results measured by answer F1.}
\label{tab:component-ablation}
\end{table}

\subsection{Analysis and Ablation}

Figure~\ref{fig:support-position-heatmap} explains where the support gains come from. The left panel shows that \method\ does not simply retrieve more candidates; it moves more gold supporting facts into the visible top-5 positions. For example, compared with NV-Embed-v2, \method\ reduces the Out mass from 31.5\% to 27.1\% and shifts more support into the second and third positions. The right panel shows the same effect in plug-in settings. IRCoT + PAGE-RAG reduces Out support from 33.4\% to 28.0\%, SelfRAG + PAGE-RAG reduces it from 15.9\% to 4.7\%, and HippoRAG2 + PAGE-RAG also yields a smaller but consistent shift. This ranking evidence supports the central mechanism: PAGE-RAG raises support-bearing paths within the candidate ranking so that they enter the fixed reader budget.

Table~\ref{tab:component-ablation} further tests the two main components behind this promotion. Removing support-aware scoring drops the average Ans F1 to 50.64, showing that graph connectivity alone is a weak signal for answer support. Removing minimal selection also hurts performance, with an average Ans F1 of 54.44, because high-scoring paths still need to be filtered into a compact and non-distracting context. The full \method\ recovers substantially higher answer quality across all three datasets, confirming that support-aware scoring and minimal selection are both necessary for the final prediction gains. Thus, the ablation verifies component necessity, while the heatmap illustrates how PAGE-RAG improves the ranking of support-bearing candidates before reading.

\section{Conclusion}

We introduced \method, a provenance-aware evidence promotion framework for multi-hop RAG. The key idea is to separate candidate reachability from final-context usefulness: expanded retrieval makes more facts available, but graph connectivity does not by itself imply answer support. PAGE-RAG builds a query-local support graph, scores candidate paths with source-tracing and support signals, and selects a compact context before generation. Experiments across sentence-level and mixed-granularity protocols show that PAGE-RAG improves both complete retrieval-to-reading pipelines and plug-in settings over existing RAG and GraphRAG backends. The ranking-position analysis further shows that PAGE-RAG promotes more gold supporting facts into the final visible context, while the component ablation confirms that both support-aware scoring and minimal selection are necessary for this improvement. More broadly, the results suggest that multi-hop RAG should not be framed only as a problem of retrieving more candidates or building more connections; it also requires deciding which connected candidates jointly support the answer. A limitation is that PAGE-RAG cannot recover facts that never enter the expanded candidate pool, so future work should study tighter integration between support promotion and adaptive candidate generation.

\section{Code Availability}

The original PAGE-RAG implementation is available at \url{https://github.com/denghk666/PAGE-RAG} under the Apache License 2.0. Third-party baseline components retain their original licenses.
\bibliography{references}

@inproceedings{guu2020realm,
  title={REALM: Retrieval-Augmented Language Model Pre-Training},
  author={Guu, Kelvin and Lee, Kenton and Tung, Zora and Pasupat, Panupong and Chang, Ming-Wei},
  booktitle={ICML},
  year={2020}
}

@inproceedings{asai2024selfrag,
  title={Self-RAG: Learning to Retrieve, Generate, and Critique through Self-Reflection},
  author={Asai, Akari and Wu, Zeqiu and Wang, Yizhong and Sil, Avirup and Hajishirzi, Hannaneh},
  booktitle={ICLR},
  year={2024}
}

@article{yan2024crag,
  title={Corrective Retrieval Augmented Generation},
  author={Yan, Shi-Qi and Gu, Jia-Chen and Zhu, Yun and Ling, Zhen-Hua},
  journal={arXiv preprint arXiv:2401.15884},
  year={2024}
}

@article{edge2024graphrag,
  title={From Local to Global: A Graph RAG Approach to Query-Focused Summarization},
  author={Edge, Darren and Trinh, Ha and Cheng, Newman and Bradley, Joshua and Chao, Alex and Mody, Apurva and Truitt, Steven and Larson, Jonathan},
  journal={arXiv preprint arXiv:2404.16130},
  year={2024}
}

@article{gutierrez2024hipporag,
  title={HippoRAG: Neurobiologically Inspired Long-Term Memory for Large Language Models},
  author={Guti{\'e}rrez, Bernal Jim{\'e}nez and Shu, Yiheng and Gu, Yu and Yasunaga, Michihiro and Su, Yu},
  journal={NeurIPS},
  year={2024}
}

@inproceedings{yang2018hotpotqa,
  title={HotpotQA: A Dataset for Diverse, Explainable Multi-hop Question Answering},
  author={Yang, Zhilin and Qi, Peng and Zhang, Saizheng and Bengio, Yoshua and Cohen, William W. and Salakhutdinov, Ruslan and Manning, Christopher D.},
  booktitle={EMNLP},
  year={2018}
}

@article{trivedi2022musique,
  title={MuSiQue: Multihop Questions via Single-hop Question Composition},
  author={Trivedi, Harsh and Balasubramanian, Niranjan and Khot, Tushar and Sabharwal, Ashish},
  journal={TACL},
  year={2022}
}

@inproceedings{karpukhin2020dpr,
  title={Dense Passage Retrieval for Open-Domain Question Answering},
  author={Karpukhin, Vladimir and Oguz, Barlas and Min, Sewon and Lewis, Patrick and Wu, Ledell and Edunov, Sergey and Chen, Danqi and Yih, Wen-tau},
  booktitle={Proceedings of the 2020 Conference on Empirical Methods in Natural Language Processing},
  year={2020}
}

@inproceedings{asai2020reasoningpaths,
  title={Learning to Retrieve Reasoning Paths over Wikipedia Graph for Question Answering},
  author={Asai, Akari and Hashimoto, Kazuma and Hajishirzi, Hannaneh and Socher, Richard and Xiong, Caiming},
  booktitle={International Conference on Learning Representations},
  year={2020}
}

@inproceedings{lewis2020rag,
  title={Retrieval-Augmented Generation for Knowledge-Intensive NLP Tasks},
  author={Lewis, Patrick and Perez, Ethan and Piktus, Aleksandra and Petroni, Fabio and Karpukhin, Vladimir and Goyal, Naman and K{\"u}ttler, Heinrich and Lewis, Mike and Yih, Wen-tau and Rockt{\"a}schel, Tim and others},
  booktitle={NeurIPS},
  year={2020}
}

@inproceedings{izacard2021fid,
  title={Leveraging Passage Retrieval with Generative Models for Open Domain Question Answering},
  author={Izacard, Gautier and Grave, Edouard},
  booktitle={EACL},
  year={2021}
}

@article{izacard2022atlas,
  title={Atlas: Few-shot Learning with Retrieval Augmented Language Models},
  author={Izacard, Gautier and Lewis, Patrick and Lomeli, Maria and Hosseini, Lucas and Petroni, Fabio and Schick, Timo and Dwivedi-Yu, Jane and Joulin, Armand and Riedel, Sebastian and Grave, Edouard},
  journal={arXiv preprint arXiv:2208.03299},
  year={2022}
}

@inproceedings{ho2020twoWiki,
  title={Constructing A Multi-hop QA Dataset for Comprehensive Evaluation of Reasoning Steps},
  author={Ho, Xanh and Duong Nguyen, Anh-Khoa and Sugawara, Saku and Aizawa, Akiko},
  booktitle={COLING},
  year={2020}
}

@article{izacard2022contriever,
  title={Unsupervised Dense Information Retrieval with Contrastive Learning},
  author={Izacard, Gautier and Caron, Mathilde and Hosseini, Lucas and Riedel, Sebastian and Bojanowski, Piotr and Joulin, Armand and Grave, Edouard},
  journal={Transactions on Machine Learning Research},
  year={2022}
}

@inproceedings{santhanam2022colbertv2,
  title={{ColBERT}v2: Effective and Efficient Retrieval via Lightweight Late Interaction},
  author={Santhanam, Keshav and Khattab, Omar and Saad-Falcon, Jon and Potts, Christopher and Zaharia, Matei},
  booktitle={Proceedings of the 2022 Conference of the North American Chapter of the Association for Computational Linguistics: Human Language Technologies},
  pages={3715--3734},
  year={2022},
  doi={10.18653/v1/2022.naacl-main.272}
}

@article{lee2024nvembed,
  title={{NV}-Embed: Improved Techniques for Training {LLM}s as Generalist Embedding Models},
  author={Lee, Chankyu and Roy, Rajarshi and Xu, Mengyao and Raiman, Jonathan and Shoeybi, Mohammad and Catanzaro, Bryan and Ping, Wei},
  journal={arXiv preprint arXiv:2405.17428},
  year={2024}
}

@inproceedings{trivedi2023ircot,
  title={Interleaving Retrieval with Chain-of-Thought Reasoning for Knowledge-Intensive Multi-Step Questions},
  author={Trivedi, Harsh and Balasubramanian, Niranjan and Khot, Tushar and Sabharwal, Ashish},
  booktitle={Proceedings of the 61st Annual Meeting of the Association for Computational Linguistics (Volume 1: Long Papers)},
  pages={10014--10037},
  year={2023},
  doi={10.18653/v1/2023.acl-long.557}
}

@article{xu2023recomp,
  title={{RECOMP}: Improving Retrieval-Augmented {LM}s with Compression and Selective Augmentation},
  author={Xu, Fangyuan and Shi, Weijia and Choi, Eunsol},
  journal={arXiv preprint arXiv:2310.04408},
  year={2023}
}

@inproceedings{hwang2025exit,
  title={{EXIT}: Context-Aware Extractive Compression for Enhancing Retrieval-Augmented Generation},
  author={Hwang, Taeho and Cho, Sukmin and Jeong, Soyeong and Song, Hoyun and Han, SeungYoon and Park, Jong C.},
  booktitle={Findings of the Association for Computational Linguistics: ACL 2025},
  pages={4895--4924},
  year={2025},
  doi={10.18653/v1/2025.findings-acl.253}
}

@inproceedings{zhuang2024efficientrag,
  title={{EfficientRAG}: Efficient Retriever for Multi-Hop Question Answering},
  author={Zhuang, Ziyuan and Zhang, Zhiyang and Cheng, Sitao and Yang, Fangkai and Liu, Jia and Huang, Shujian and Lin, Qingwei and Rajmohan, Saravan and Zhang, Dongmei and Zhang, Qi},
  booktitle={Proceedings of the 2024 Conference on Empirical Methods in Natural Language Processing},
  pages={3392--3411},
  year={2024},
  doi={10.18653/v1/2024.emnlp-main.199}
}

@article{liu2026opera,
  author  = {Liu, Yu and Liu, Yanbing and Yuan, Fangfang and Cao, Cong and Sun, Youbang and Peng, Kun and Chen, WeiZhuo and Li, Jianjun and Ma, Zhiyuan},
  title   = {{OPERA}: A Reinforcement Learning--Enhanced Orchestrated Planner-Executor Architecture for Reasoning-Oriented Multi-Hop Retrieval},
  journal = {Proceedings of the AAAI Conference on Artificial Intelligence},
  volume  = {40},
  number  = {38},
  pages   = {32258--32266},
  year    = {2026},
  doi     = {10.1609/aaai.v40i38.40499}
}

@article{gutierrez2025hipporag2,
  title={From {RAG} to Memory: Non-Parametric Continual Learning for Large Language Models},
  author={Guti{\'e}rrez, Bernal Jim{\'e}nez and Shu, Yiheng and Qi, Weijian and Zhou, Sizhe and Su, Yu},
  journal={arXiv preprint arXiv:2502.14802},
  year={2025}
}

@inproceedings{guo2025lightrag,
  title={{LightRAG}: Simple and Fast Retrieval-Augmented Generation},
  author={Guo, Zirui and Xia, Lianghao and Yu, Yanhua and Ao, Tu and Huang, Chao},
  booktitle={Findings of the Association for Computational Linguistics: EMNLP 2025},
  pages={10746--10761},
  year={2025},
  doi={10.18653/v1/2025.findings-emnlp.568}
}

@inproceedings{zhu2025kg2rag,
  title={Knowledge Graph-Guided Retrieval Augmented Generation},
  author={Zhu, Xiangrong and Xie, Yuexiang and Liu, Yi and Li, Yaliang and Hu, Wei},
  booktitle={Proceedings of the 2025 Conference of the Nations of the Americas Chapter of the Association for Computational Linguistics: Human Language Technologies (Volume 1: Long Papers)},
  pages={8912--8924},
  year={2025},
  doi={10.18653/v1/2025.naacl-long.449}
}

@article{luo2025gfmrag,
  title={{GFM}-{RAG}: Graph Foundation Model for Retrieval Augmented Generation},
  author={Luo, Linhao and Zhao, Zicheng and Haffari, Gholamreza and Phung, Dinh and Gong, Chen and Pan, Shirui},
  journal={arXiv preprint arXiv:2502.01113},
  year={2025}
}

@inproceedings{yao2023react,
  title={{ReAct}: Synergizing Reasoning and Acting in Language Models},
  author={Yao, Shunyu and Zhao, Jeffrey and Yu, Dian and Du, Nan and Shafran, Izhak and Narasimhan, Karthik and Cao, Yuan},
  booktitle={International Conference on Learning Representations},
  year={2023}
}

@inproceedings{jiang2024longllmlingua,
  title={{LongLLMLingua}: Accelerating and Enhancing {LLM}s in Long Context Scenarios via Prompt Compression},
  author={Jiang, Huiqiang and Wu, Qianhui and Luo, Xufang and Li, Dongsheng and Lin, Chin-Yew and Yang, Yuqing and Qiu, Lili},
  booktitle={Proceedings of the 62nd Annual Meeting of the Association for Computational Linguistics (Volume 1: Long Papers)},
  pages={1658--1677},
  year={2024},
  doi={10.18653/v1/2024.acl-long.91}
}

@article{deepseek2026v4,
  title={{DeepSeek-V4}: Towards Highly Efficient Million-Token Context Intelligence},
  author={{DeepSeek-AI}},
  journal={arXiv preprint arXiv:2606.19348},
  year={2026}
}

\appendix

\section{Appendix A: Method Details}

This appendix provides the implementation details behind the PAGE-RAG pipeline. The main paper describes the high-level workflow: retrieve an expanded candidate pool, build a query-local graph, score support-bearing paths, and select a fixed-budget reader context. Here we make the graph construction, scoring signals, and minimal selection procedure explicit.

\begin{table}[h]
\centering
\small
\begin{tabular}{lll}
\toprule
\textbf{Edge type} & \textbf{Endpoints} & \textbf{Role} \\
\midrule
contains & document--sentence & Source anchoring \\
mention & sentence--entity & Entity anchoring \\
adjacent & sentence--sentence & Local context link \\
shared entity & sentence--sentence & Cross-source bridge \\
relation & entity--entity & Optional relation link \\
\bottomrule
\end{tabular}
\caption{Graph edge types used in the query-local PAGE-RAG graph.}
\label{tab:appendix-edge-types}
\end{table}

\subsection{A.1 Query-Local Graph Construction}

PAGE-RAG builds a graph only over the candidates retrieved for the current question. It does not require a corpus-wide graph index. Each graph contains three node types: document or passage nodes, sentence nodes, and entity nodes. The graph also contains five edge types, summarized in Table~\ref{tab:appendix-edge-types}. Document--sentence and sentence--entity edges keep source traceability, while sentence--sentence edges are the main paths used for support promotion.

The construction resembles graph-based RAG systems that organize text through passages, entities, and relation links \citep{edge2024graphrag,gutierrez2024hipporag,guo2025lightrag,zhu2025kg2rag}, but PAGE-RAG uses the graph differently. A graph edge is treated as a proposal for support, not as support by default. This is the implementation-level form of the connectivity-support gap discussed in the main paper.

\subsection{A.2 Feature Notation}

We list the feature calculations used by the implementation. \(\operatorname{Clamp}(x)\) clips \(x\) into \([0,1]\). \(\operatorname{Tok}(x)\) denotes lowercase, stopword-removed tokens. For two text fields \(a\) and \(b\), lexical coverage is:
\begin{equation}
\label{eq:appendix-basic-lex}
\operatorname{Lex}(a,b)=
\frac{\sum_t \min(\operatorname{count}_a(t),\operatorname{count}_b(t))}
{\max(|\operatorname{Tok}(a)|,1)}.
\end{equation}
For two sets \(A\) and \(B\), PAGE-RAG uses \(\operatorname{Jaccard}(A,B)=|A\cap B|/|A\cup B|\).

The sentence-level relevance used for seed selection and path scoring combines lexical matching with question-entity coverage:
\begin{equation}
\label{eq:appendix-sentence-relevance}
\begin{aligned}
\operatorname{sent\_rel}(q,s)
&=\operatorname{Clamp}\!\big(
w_{\ell}\operatorname{Lex}(q,s)\\
&\quad+w_{e}\operatorname{ent\_overlap}(q,s)\big),\\
\operatorname{ent\_overlap}(q,s)
&=\frac{|\operatorname{Ent}(q)\cap \operatorname{Ent}(s)|}
{\max(|\operatorname{Ent}(q)|,1)}.
\end{aligned}
\end{equation}
Seed scoring further adds a soft relation-intent boost and a bounded literal-entity bonus:
\begin{equation}
\label{eq:appendix-seed-score}
\begin{aligned}
\operatorname{seed}(q,s)
&=\operatorname{sent\_rel}(q,s)
+\operatorname{rel\_boost}(q,s)\\
&\quad+\operatorname{literal}(q,s),\\
\operatorname{literal}(q,s)
&=\min(\tau_{\mathrm{lit}},\lambda_{\mathrm{lit}}n_{\mathrm{lit}}(q,s)),
\end{aligned}
\end{equation}
where \(n_{\mathrm{lit}}(q,s)\) is the number of question entities that appear literally in \(s\).

\subsection{A.3 Metadata Feature Calculation}

For every edge \(e\), PAGE-RAG stores source-tracing metadata:
\begin{equation}
\label{eq:appendix-metadata}
m(e)=\{\mathrm{src},\mathrm{sent},\mathrm{trig},\mathrm{conf},
\mathrm{div},\mathrm{spec},\mathrm{hub}\}.
\end{equation}
Here, \(\mathrm{src}\) and \(\mathrm{sent}\) identify the original source and sentence, and \(\mathrm{trig}\) is the entity or relation that creates the edge. For an edge \(e\), \(T_e\) denotes its trigger set; for a shared-entity edge between sentences \(s_i\) and \(s_j\), \(T_e=\operatorname{Ent}(s_i)\cap\operatorname{Ent}(s_j)\). \(D_e\) denotes the set of distinct source documents in the edge provenance, and \(T\) denotes all trigger entities in the query-local graph. The remaining fields are computed as follows.

\textbf{Confidence.}
For contains, mention, and adjacent edges, \(\operatorname{conf}(e)=1.0\). For relation edges, \(\operatorname{conf}(e)\) is the extracted relation confidence. For shared-entity edges with trigger set \(T_e\):
\begin{equation}
\label{eq:appendix-confidence}
\operatorname{conf}(e)=\min(1.0,b_0+b_1|T_e|).
\end{equation}

\textbf{Source diversity.}
PAGE-RAG computes source diversity with a conservative scaling so that a single-source edge receives no diversity reward:
\begin{equation}
\label{eq:appendix-diversity}
\operatorname{div}(e)=
\begin{cases}
0, & |D_e|=0,\\
\min(1.0,\rho(|D_e|-1)), & \text{otherwise}.
\end{cases}
\end{equation}

\textbf{Specificity and hubness.}
For an entity \(t\), \(\operatorname{df}(t)\) is the number of sentences in the query-local graph that mention \(t\), and:
\begin{equation}
\label{eq:appendix-idf}
\operatorname{idf}(t)=
\log\frac{1+|\mathcal{S}_q|}{1+\operatorname{df}(t)}+1,
\end{equation}
where \(\mathcal{S}_q\) is the sentence set in the query-local graph. Adjacent and non-shared edges use fixed fallback values because they do not have a shared trigger set. For shared-entity edges:
\begin{equation}
\label{eq:appendix-spec-hub}
\begin{aligned}
\operatorname{spec}(e)&=\operatorname{Clamp}\!\left(
\frac{\operatorname{mean}_{t\in T_e}\operatorname{idf}(t)}
{\max_{t'\in T}\operatorname{idf}(t')}\right),\\
\operatorname{hub}(e)&=\operatorname{Clamp}\!\left(
\frac{\operatorname{mean}_{t\in T_e}\operatorname{df}(t)}
{\max_{t'\in T}\operatorname{df}(t')}\right).
\end{aligned}
\end{equation}

\textbf{Noise penalty.}
For a shared-entity edge, PAGE-RAG accumulates a bounded noise penalty when the edge has low specificity, low confidence, or weak source diversity:
\begin{equation}
\label{eq:appendix-noise}
\begin{aligned}
\widetilde{\operatorname{noise}}(e)
&=\xi_s\mathbb{I}[\operatorname{spec}(e)<\tau_s]
+\xi_c\mathbb{I}[\operatorname{conf}(e)<\tau_c]\\
&\quad+\xi_d\mathbb{I}[|D_e|=1\ \mathrm{and}\ \neg\mathrm{paragraph}],\\
\operatorname{noise}(e)
&=\operatorname{Clamp}(\widetilde{\operatorname{noise}}(e)).
\end{aligned}
\end{equation}
Adjacent edges use a small fallback noise penalty.

\subsection{A.4 Edge and Path Feature Calculation}

The edge relevance feature uses the question tokens, the provenance sentences attached to the edge, and the trigger text. Here \(\operatorname{prov}(e)\) denotes the concatenated provenance sentences for \(e\), and \(\operatorname{trig}(e)\) denotes its trigger entity or relation:
\begin{equation}
\label{eq:appendix-edge-relevance}
\operatorname{edge\_rel}(e,q)=
\operatorname{Jaccard}\!\left(\operatorname{Tok}(q),
\operatorname{Tok}(\operatorname{prov}(e)+\operatorname{trig}(e))\right).
\end{equation}
The reliability feature combines diversity and confidence:
\begin{equation}
\label{eq:appendix-reliability}
\operatorname{reliability}(e)=
\operatorname{Clamp}\!\left(\lambda_d\operatorname{div}(e)+\lambda_c\operatorname{conf}(e)\right).
\end{equation}
The implementation then forms an edge score from these feature values:
\begin{equation}
\label{eq:appendix-edge-feature-score}
\begin{aligned}
\operatorname{edge}(e,q)=
&\alpha \operatorname{edge\_rel}(e,q)
+\beta \operatorname{reliability}(e)\\
&+\gamma \operatorname{spec}(e)
-\delta \operatorname{hub}(e)
-\eta \operatorname{noise}(e).
\end{aligned}
\end{equation}
The edge score linearly combines relevance, reliability, specificity, hubness, and noise terms. For a shared-entity edge whose trigger appears in the question or question entities, PAGE-RAG can add a small query-entity edge boost and discount the hub penalty.

For a candidate path \(\pi\), let \(\mathcal{E}_\pi\) be the edge set along the path. PAGE-RAG first computes a sentence base. For one-sentence paths it uses the maximum sentence relevance; for multi-sentence paths it uses the mean sentence relevance:
\begin{equation}
\label{eq:appendix-sentence-base}
\operatorname{sent\_base}(\pi)=
\begin{cases}
\max_{s\in \pi}\operatorname{sent\_rel}(q,s), & |\pi|=1,\\
\operatorname{mean}_{s\in \pi}\operatorname{sent\_rel}(q,s), & |\pi|>1.
\end{cases}
\end{equation}
For neighboring sentence pairs \((s_i,s_{i+1})\) in a path, coherence is:
\begin{equation}
\label{eq:appendix-coherence}
\begin{aligned}
\operatorname{pair\_coh}(s_i,s_{i+1})
&=\omega_e\operatorname{Jaccard}(\operatorname{Ent}(s_i),\operatorname{Ent}(s_{i+1}))\\
&\quad+\omega_{\ell}\operatorname{Lex}(s_i,s_{i+1}),\\
\operatorname{coh}(\pi)&=\operatorname{mean}_i\operatorname{pair\_coh}(s_i,s_{i+1}).
\end{aligned}
\end{equation}
In paragraph mode, \(\operatorname{coh}(\pi)=0\), because cross-paragraph lexical coherence can double-count hub links.

The path length term is:
\begin{equation}
\label{eq:appendix-length-cost}
\operatorname{len\_cost}(\pi)=
\begin{cases}
\mu\max(0,|\mathcal{E}_\pi|-1), & \text{standard mode},\\
-\mu_p\min(2,|\mathcal{E}_\pi|), & \text{paragraph mode}.
\end{cases}
\end{equation}
The second case is a small multi-hop bonus, since paragraph-style corpora often require crossing paragraph boundaries. The path score used for ranking is:
\begin{equation}
\label{eq:appendix-path-feature-score}
\begin{aligned}
\operatorname{path}(\pi,q)=
\operatorname{sent\_base}(\pi)
+\sum_{e\in \mathcal{E}_\pi}\operatorname{edge}(e,q)\\
+\lambda_{\mathrm{coh}}\operatorname{coh}(\pi)
-\operatorname{len\_cost}(\pi).
\end{aligned}
\end{equation}
For bridge questions, a path receives an additional bounded bonus when it contains at least two sentences, one sentence mentions a question entity, and another sentence carries a relation signal. The contains and mention edges mainly provide traceability and entity anchors; path ranking is computed over sentence-level transitions induced by adjacent, shared-entity, and relation or semantic links.

\subsection{A.5 Minimal Sufficient Selection}

After path scoring, PAGE-RAG must convert a ranked path list into a compact reader context. The selector is deliberately different from simply taking the top-scoring sentences. It adds paths only when they improve the current support state or provide a plausible intermediate bridge toward a sufficient context. This design is important because expanded retrieval increases reachability and noise at the same time. The concrete selection flow is shown in Algorithm~\ref{alg:appendix-minimal-selection}.

\setcounter{table}{5}
\begin{table*}[!b]
\centering
\small
\resizebox{\textwidth}{!}{
\begin{tabular}{lrrrrrrrr}
\toprule
\textbf{Method}
& \multicolumn{2}{c}{HotpotQA}
& \multicolumn{2}{c}{MuSiQue}
& \multicolumn{2}{c}{2Wiki}
& \multicolumn{2}{c}{Avg.} \\
\cmidrule(lr){2-3}\cmidrule(lr){4-5}\cmidrule(lr){6-7}\cmidrule(lr){8-9}
& Ans F1 & Sup F1 & Ans F1 & Sup F1 & Ans F1 & Sup F1 & Ans F1 & Sup F1 \\
\midrule
w/o query alignment & 56.10 & 27.79 & 43.20 & 46.41 & 44.78 & 29.41 & 48.35 & 30.71 \\
w/o provenance reliability & 62.94 & 40.60 & 39.44 & 40.14 & 49.38 & 36.17 & 52.79 & 38.06 \\
w/o bridge specificity & 61.66 & 39.40 & 36.60 & 36.27 & 48.81 & 35.53 & 51.74 & 36.89 \\
w/o path quality control & 63.08 & 40.08 & 39.90 & 40.39 & 48.15 & 34.56 & 52.20 & 37.01 \\
\midrule
PAGE-RAG & 70.94 & 55.09 & 56.00 & 62.00 & 63.23 & 54.16 & 65.00 & 55.31 \\
\bottomrule
\end{tabular}
}
\caption{Feature-group ablations. Avg. is weighted by the number of evaluated examples.}
\label{tab:appendix-feature-ablation}
\end{table*}
\setcounter{table}{4}

\begin{algorithm}[t]
\caption{Minimal sufficient selection}
\label{alg:appendix-minimal-selection}
\begin{algorithmic}[1]
\REQUIRE Ranked candidate paths \(P\), budget \(k\), question \(q\)
\ENSURE Selected context \(C\)
\STATE \(C \leftarrow \emptyset\), state \(\leftarrow\) Ambiguous
\FOR{each path \(\pi\in P\)}
    \STATE \(C' \leftarrow C \cup \mathrm{sentences}(\pi)\)
    \IF{\(|C'| > k\)}
        \STATE skip \(\pi\)
    \ELSIF{\(\pi\) improves support utility or state}
        \STATE \(C \leftarrow C'\)
        \STATE update state from selected context
    \ELSIF{\(\pi\) is a plausible intermediate bridge}
        \STATE keep one-step patience before rejecting
    \ELSE
        \STATE reject \(\pi\)
    \ENDIF
    \IF{state is Enough}
        \STATE break
    \ENDIF
\ENDFOR
\STATE fill remaining budget, if needed, with highest-scoring unused candidates
\RETURN \(C\)
\end{algorithmic}
\end{algorithm}

The last fill step is used only to keep the reader budget fixed in controlled experiments. It first preserves the minimal selected core and then fills unused slots with the highest-scoring remaining candidates from the expanded pool. This prevents an unfair comparison where PAGE-RAG receives fewer final context units than the baseline.

\subsection{A.6 Document and Chunk-Level Adaptation}

The main PAGE-RAG pipeline operates at sentence level, but the same principle can be applied to document, passage, or chunk-level backends. In mixed-granularity settings, PAGE-RAG still builds internal sentence-level or chunk-internal paths when text structure is available, because sentence-level links make support scoring more precise. The final output, however, is mapped back to the backend's native unit, such as a document or chunk identifier. Edge metadata are also interpreted at the native granularity: source diversity becomes diversity across documents or chunks, and support paths are promoted only when their selected units remain within the final top-\(k\) budget.

This adaptation is why PAGE-RAG can serve as a plug-in layer. The upstream backend is responsible for producing a candidate pool in its own format; PAGE-RAG uses a temporary graph to recalibrate the pool by support and returns the same type of context unit expected by the downstream reader.

\section{Appendix B: Feature-Group Ablation}

\begin{table}[t]
\centering
\small
\begin{tabular}{lp{0.57\columnwidth}}
\toprule
\textbf{Group} & \textbf{Removed signals} \\
\midrule
Query alignment & sentence relevance, edge relevance, question entity overlap, relation-type boost \\
Provenance reliability & confidence, source diversity, provenance reliability score \\
Bridge specificity & entity specificity, hub penalty, query-entity hub discount \\
Path quality control & noise penalty, path coherence, path length cost or bonus \\
\bottomrule
\end{tabular}
\caption{Feature groups removed in Appendix B ablations.}
\label{tab:appendix-feature-groups}
\end{table}
\setcounter{table}{6}

The main paper reports component-level ablations for support-aware scoring and minimal selection. Here we further ask whether the scoring signals are merely a collection of features or whether each semantic group contributes to support promotion. We group the support scoring signals into four interpretable families.

Query alignment keeps path exploration anchored to the current question, using sentence relevance, edge relevance, question-entity overlap, and relation-intent cues. Provenance reliability estimates whether a connection is backed by reliable source traces, using confidence and source diversity. Bridge specificity addresses the connectivity-support gap directly: shared entities can create useful bridges, but generic hub entities often connect topical distractors, so specificity and hubness must be separated. Path quality control measures whether a candidate path is coherent and compact rather than merely connected, using noise, coherence, and length-related terms. Table~\ref{tab:appendix-feature-groups} summarizes which signals are removed in each ablation.

The results show that the scoring module is not driven by a single signal, and that different datasets stress different parts of the scorer. This is expected because the three benchmarks differ in how the missing support usually appears in the expanded pool: HotpotQA often contains explicit bridge entities, MuSiQue contains paragraph-level semantic associations, and 2Wiki contains relation chains over named entities.

On HotpotQA and 2Wiki, removing query alignment causes the largest drop. HotpotQA often requires linking an entity-bearing bridge sentence to an answer-bearing sentence, and 2Wiki frequently asks relation-chain questions over named entities. In these settings, graph connectivity alone can easily follow a related entity neighborhood without staying anchored to the exact question. Removing query alignment drops HotpotQA from 70.94 to 56.10 Ans F1 and drops 2Wiki from 63.23 to 44.78 Ans F1; the support drops are also large.

MuSiQue behaves differently. Its questions are paragraph-style and often require a more implicit decomposition, so surface query overlap is less dominant. The largest answer drop occurs when bridge specificity is removed, and path quality control is also important. This matches the dataset structure: many candidates can be loosely connected through generic entities or long paragraph-level associations, but only a subset forms a compact chain that supports the answer. Specificity and path quality therefore help PAGE-RAG distinguish useful bridges from plausible but distracting ones.

Provenance reliability gives a steadier contribution across datasets. It does not always produce the single largest drop, but it consistently improves both answer and support quality by separating edges with clearer source traces from weaker or less repeated links. This behavior is useful for fixed-budget selection: when two candidate paths have similar topical relevance, the path with clearer source traces is less likely to be a coincidental connection. The result is not that provenance alone solves support promotion, but that it stabilizes the scorer when relevance and graph structure are ambiguous.

The four groups are also complementary. Query alignment tells PAGE-RAG what relation the current question is asking for; bridge specificity and path quality decide whether a connected path is a plausible support path rather than a hub-induced shortcut; provenance reliability checks whether the connection has credible source traces. Removing any one of them leaves the model with a partial view of support. Overall, the full scorer is substantially stronger than any feature-group removal, supporting the design choice to treat connectivity as a hypothesis that must be calibrated by all four signal families together.

\setcounter{table}{7}
\begin{table*}[!b]
\centering
\small
\resizebox{\textwidth}{!}{
\begin{tabular}{p{0.09\textwidth}p{0.37\textwidth}p{0.15\textwidth}p{0.15\textwidth}p{0.16\textwidth}}
\toprule
\textbf{Dataset} & \textbf{Question} & \textbf{Gold answer} & \textbf{Initial answer} & \textbf{PAGE-RAG answer} \\
\midrule
HotpotQA & Which team's 2013-2014 season had players including a Slovenian who plays at both point guard and shooting guard? & Phoenix Suns & Slovenia & Phoenix Suns \\
MuSiQue & When did the first mosque open in the place where Tobolar Copra plant is located? & September 2012 & 1970 & September 2012 \\
2Wiki & Who is Ibrahim Shah of Selangor's paternal grandfather? & Daeng Chelak & Sultan Salehuddin Shah & Almarhum Daeng Chelak \\
\bottomrule
\end{tabular}
}
\caption{Case study summary. Initial answer denotes the answer produced from the initial retrieval context, while PAGE-RAG answer denotes the answer after support promotion.}
\label{tab:appendix-case-metrics}
\end{table*}
\setcounter{table}{6}

\section{Appendix C: Noise Robustness}

This appendix evaluates whether PAGE-RAG remains stable when the expanded candidate pool becomes noisier. The main experiments already use an expanded top-20 retrieval pool and a fixed top-5 reader context. In this stress test, we further append additional distractor documents to the expanded pool before graph construction. The added distractors are sampled from the same benchmark corpus but are not part of the original top-20 cache for the current question. PAGE-RAG then builds the query-local graph over this noisier pool, performs the same support-aware path scoring and minimal selection procedure, and still returns only five context units to the reader.

We test three noise levels: \(+5\), \(+10\), and \(+20\) extra distractor documents. All runs use the same DeepSeek-V4-Pro reader as the main table and keep the same final reader budget. Table~\ref{tab:appendix-noise-robustness} reports answer F1.

The results show a gradual degradation rather than a collapse. On HotpotQA, adding five, ten, and twenty extra distractors reduces answer F1 by 2.68, 4.07, and 5.26 points. On MuSiQue, the corresponding drops are 4.93, 5.43, and 6.62 points. On 2Wiki, the \(+5\), \(+10\), and \(+20\) settings reduce answer F1 by 5.65, 6.97, and 7.86 points. Compared with the main-table baselines, these noisy PAGE-RAG variants remain stronger than, or roughly comparable to, common strong baselines under the same final budget, which indicates that the method keeps substantial answer quality even when the expanded pool is polluted. At the same time, the monotonic decline is expected: when the candidate pool becomes much noisier, the query-local graph contains more plausible but non-supporting bridges, and support promotion becomes harder. We therefore interpret this experiment as a sensitivity analysis rather than a claim of noise-invariant robustness.

\begin{center}
\centering
\small
\begin{tabular}{lrrrr}
\toprule
\textbf{Dataset} & \textbf{PAGE-RAG} & \textbf{+5} & \textbf{+10} & \textbf{+20} \\
\midrule
HotpotQA & 70.94 & 68.26 & 66.87 & 65.68 \\
MuSiQue & 56.00 & 51.07 & 50.57 & 49.38 \\
2Wiki & 63.23 & 57.58 & 56.26 & 55.37 \\
\bottomrule
\end{tabular}
\captionof{table}{Noise robustness measured by answer F1. Each \(+m\) column appends \(m\) extra distractor documents to the expanded candidate pool while keeping the final reader context fixed at five units.}
\label{tab:appendix-noise-robustness}
\end{center}
\setcounter{table}{8}

\section{Appendix D: Case Study}

We provide three examples where the initial top-5 retrieval misses at least one required supporting fact, while the expanded top-20 pool contains the missing fact and PAGE-RAG promotes it into the final top-5 reader context. Table~\ref{tab:appendix-case-metrics} summarizes the questions and answer changes, and Table~\ref{tab:appendix-case-facts} shows how the missing facts move from the expanded pool into the final context.

\textbf{HotpotQA.}
The initial retrieval is dominated by sentences about basketball positions. It retrieves that Goran Dragic can play both point guard and shooting guard, but it does not include the facts needed to connect him to the 2013-2014 Phoenix Suns season. PAGE-RAG promotes the nationality fact and two season facts from ranks 6, 8, and 12 into the final context. As a result, the reader changes its answer from ``Slovenia'' to ``Phoenix Suns''.

\textbf{MuSiQue.}
The initial top-5 retrieval is attracted by the phrase ``first mosque'' and returns mosque-related distractors from Australia, the United Kingdom, Ukraine, and Cambodia. The expanded pool contains the missing location chain: Tobolar Copra is in Majuro, and Majuro's first mosque opened in September 2012. PAGE-RAG promotes these two facts from ranks 13 and 15 into the final context, changing the answer from ``1970'' to the correct date.

\textbf{2Wiki.}
The initial top-5 retrieval contains topically related Selangor fragments but misses the complete paternal chain. The expanded pool contains both missing links: Ibrahim Shah was born Raja Ibrahim bin Raja Lumu, and Salehuddin Shah was born Raja Lumu bin Daeng Chelak. PAGE-RAG promotes these facts from ranks 6 and 19 into the first two final positions, changing the answer from Ibrahim's father to the correct grandfather.

Across the three examples, PAGE-RAG helps for the same reason: the expanded pool already contains the missing hop, but it is buried behind topical distractors. By scoring connected candidates as support hypotheses and selecting a compact set of complementary paths, PAGE-RAG moves the missing facts into the final reader context without increasing the reader budget.

\begin{table*}[!b]
\centering
\small
\resizebox{\textwidth}{!}{
\begin{tabular}{p{0.09\textwidth}p{0.36\textwidth}p{0.10\textwidth}p{0.12\textwidth}p{0.25\textwidth}}
\toprule
\textbf{Dataset} & \textbf{Promoted fact} & \textbf{Initial rank} & \textbf{Final rank} & \textbf{Role in the answer chain} \\
\midrule
HotpotQA & Goran Dragic is a Slovenian professional basketball player. & Top-6 & Final-2 & identifies the Slovenian player \\
 & The 2013-14 Phoenix Suns season was the team's 46th NBA season. & Top-12 & Final-3 & identifies the team season \\
 & Goran Dragic was a returning player for the Suns in that season. & Top-8 & Final-5 & links the player to the season \\
\midrule
MuSiQue & Tobolar Copra processing plant is in Majuro, Marshall Islands. & Top-13 & Final-2 & resolves the location in the question \\
 & The first mosque in Majuro opened in September 2012. & Top-15 & Final-3 & provides the requested date \\
\midrule
2Wiki & Ibrahim Shah was born Raja Ibrahim bin Raja Lumu. & Top-6 & Final-1 & links Ibrahim Shah to Raja Lumu \\
 & Salehuddin Shah was born Raja Lumu bin Daeng Chelak. & Top-19 & Final-2 & links Raja Lumu to Daeng Chelak \\
\bottomrule
\end{tabular}
}
\caption{Promoted supporting facts. Initial rank is the rank in the expanded top-20 retrieval pool; final rank is the rank after PAGE-RAG selection.}
\label{tab:appendix-case-facts}
\end{table*}

\end{document}